\documentclass[10pt,twocolumn,letterpaper]{article}

\usepackage[pagenumbers]{cvpr} 

\usepackage{multirow}

\definecolor{cvprblue}{rgb}{0.21,0.49,0.74}
\usepackage[pagebackref,breaklinks,colorlinks,allcolors=cvprblue]{hyperref}

\def\paperID{34387} 
\def\confName{CVPR}
\def\confYear{2026}

\title{SHAPE: Structure-aware Hierarchical Unsupervised Domain Adaptation with Plausibility Evaluation for Medical Image Segmentation}

\author{
    Linkuan Zhou\textsuperscript{1}, 
    Yinghao Xia\textsuperscript{1}, 
    Yufei Shen\textsuperscript{1}, 
    Xiangyu Li\textsuperscript{2}, 
    Wenjie Du\textsuperscript{3}, \\
    Cong Cong\textsuperscript{4}, 
    Leyi Wei\textsuperscript{5},
    Ran Su\textsuperscript{6}, 
    Qiangguo Jin\textsuperscript{1,*} \\
    \textsuperscript{1}School of Software, Northwestern Polytechnical University, \\
    \textsuperscript{2}School of Computer Science and Technology, Harbin Institute of Technology, \\ 
    \textsuperscript{3}School of Software Engineering, USTC, \\
    \textsuperscript{4}Australian Institute of Health Innovation, Macquarie University,\\
    \textsuperscript{5}Faculty of Applied Science, Macao Polytechnic University,\\
    \textsuperscript{6}School of Computer Software, Tianjin University,    \\
    {\tt\small qgking@nwpu.edu.cn}
}

\begin{document}
\maketitle
\renewcommand{\thefootnote}{\fnsymbol{footnote}} 
\footnotetext[1]{Corresponding author}
\begin{abstract}
\par Unsupervised Domain Adaptation (UDA) is essential for deploying medical segmentation models across diverse clinical environments. Existing methods are fundamentally limited, suffering from semantically unaware feature alignment that results in poor distributional fidelity and from pseudo-label validation that disregards global anatomical constraints, thus failing to prevent the formation of globally implausible structures. To address these issues, we propose SHAPE (\textbf{S}tructure-aware \textbf{H}ierarchical Unsupervised Domain \textbf{A}daptation with \textbf{P}lausibility \textbf{E}valuation), a framework that reframes adaptation towards global anatomical plausibility. Built on a DINOv3 foundation, its Hierarchical Feature Modulation (HFM) module first generates features with both high fidelity and class-awareness. 
This shifts the core challenge to robustly validating pseudo-labels. To augment conventional pixel-level validation, we introduce Hypergraph Plausibility Estimation (HPE), which leverages hypergraphs to assess the global anatomical plausibility that standard graphs cannot capture.
This is complemented by Structural Anomaly Pruning (SAP) to purge remaining artifacts via cross-view stability.
SHAPE significantly outperforms prior methods on cardiac and abdominal cross-modality benchmarks, achieving state-of-the-art average Dice scores of 90.08\% (MRI$\to$CT) and 78.51\% (CT$\to$MRI) on cardiac data, and 87.48\% (MRI$\to$CT) and 86.89\% (CT$\to$MRI) on abdominal data. The code is available at
\url{https://github.com/BioMedIA-repo/SHAPE}.
\end{abstract}    
\section{Introduction}
\label{sec:intro}

Medical image segmentation plays a crucial role in medical analysis and treatment. Despite the success of deep learning models in this task, they typically assume that training and test data follow the same distribution~\cite{ronneberger2015u}. However, this hypothesis is not applicable in real-world clinical scenarios due to the large variations in imaging equipment, modalities, and parameters. Thus, the performance of a model trained on a labeled source domain may decline sharply when applied to an unlabeled target domain~\cite{kumari2023deep}. Unsupervised Domain Adaptation (UDA) addresses this by transferring knowledge from a labeled source to an unlabeled target domain, avoiding costly re-annotation~\cite{han2021deep}.

Recently, deep learning methods have been used to address the UDA problem\cite{zeng2024reliable,zhang2024mapseg,zhang2024iplc}.
These methods can be broadly categorized into alignment based methods and pseudo label based methods. The first category aligns the source and target domains based on image appearance~\cite{zhu2017unpaired,tomar2021self,zeng2024reliable,xu2023asc}, feature distribution~\cite{tsai2018learning,dou2019pnp,chen2020unsupervised,zhao2022uda}, or output prediction consistency\cite{vu2019advent,wang2023shape}. For example, RSA~\cite{zeng2024reliable} employed a conditional diffusion model to generate multiple source-like images, and used predictive consistency to select the most reliable generated image.
In contrast, LE-UDA~\cite{zhao2022uda} focused on feature-level alignment, constructing self-ensembling consistency to facilitate knowledge transfer between domains and utilizing an adversarial learning module for UDA.
Different from the above methods, SIFA~\cite{chen2020unsupervised} performed co-alignment of domains from both image and feature perspectives, simultaneously transforming the appearance of images across domains and enhancing the domain invariance of the extracted features by leveraging adversarial learning in multiple aspects. 
In addition, SADA~\cite{wang2023shape} aligned the joint distribution of segmentation results between source and target images, thereby achieving domain adaptation. However, these methods typically learn a global, content-agnostic mapping between domains, enabling overall domain adaptation, inevitably breaks the fine-grained mapping relationships between cross-domain classes and categories.

The second category leverages semi-supervised learning to generate pseudo labels for target domain data using source domain data. For example, MAPSeg~\cite{zhang2024mapseg} introduced a masked autoencoding and pseudo labeling segmentation framework, which demonstrated good performance in heterogeneous and volumetric medical image segmentation. Similarly, 
GenericSSL~\cite{wang2024towards} proposes a knowledge distillation framework, utilizing a shared diffusion encoder to learn distribution-invariant features and a reweighted decoder to generate reliable pseudo-labels for further supervision.
In order to generate reliable pseudo labels, IPLC~\cite{zhang2024iplc} iteratively generated pseudo labels using pre-trained source models and the SAM-Med2D~\cite{cheng2023sam}, incorporating multiple random sampling and entropy estimation while continuously updating prompts for domain adaptation. Unlike IPLC, UPL-SFDA~\cite{wu2023upl} generated different predictions of the target domain by duplicating the pre-trained model’s prediction head multiple times with perturbations and generated reliable pseudo labels using uncertainty estimation. Nevertheless, their quality assessment relies heavily on local, pixel-wise metrics such as prediction entropy or consistency, potentially admitting pseudo-labels with anatomical structural flaws, thus hindering the model from learning generalizable and structurally coherent features. 

Despite progress, these approaches face two fundamental challenges that limit their efficacy. First, feature alignment often operates in a semantically unaware manner. Monolithic strategies apply a uniform transformation across the feature map, averaging style characteristics over distinct anatomical structures. This blending prevents the generation of class-specific style information, leading to imprecise feature alignment and poor distributional fidelity. Second, pseudo-label validation disregards global anatomical constraints. Existing methods rely on pixel-level confidence or local consistency and are unable to prevent the formation of pseudo-labels that constitute globally implausible structures, such as those with anatomically impossible shapes or spatial arrangements.

To overcome these limitations, we propose SHAPE (\textbf{S}tructure-aware \textbf{H}ierarchical Unsupervised Domain \textbf{A}daptation with \textbf{P}lausibility \textbf{E}valuation), a UDA framework that shifts the paradigm from local pixel correctness to global anatomical plausibility. Built upon a powerful DINOv3~\cite{simeoni2025dinov3} foundation, SHAPE introduces a synergistic pipeline. Our Hierarchical Feature Modulation (HFM) performs a class-aware, spatially-differentiated alignment by applying tailored mixing strategies to semantic cores and structural boundaries, generating features with high distributional fidelity.
To leverage these adapted features for self-training, high-quality pseudo-labels are essential. Such labels must be not only pixel-wise accurate but also globally plausible in their anatomical structure. While standard graphs capture pairwise relations, they fall short of representing the holistic interplay of multiple anatomical structures. We therefore introduce Hypergraph Plausibility Estimation (HPE), which leverages hypergraphs to uniquely model and score both the intra-class shape of individual structures and the inter-class spatial arrangement of the entire anatomy. Finally, a Structural Anomaly Pruning (SAP) stage inspects cross-view stability to identify and purge spurious, hallucinated class-level predictions, ensuring maximum label fidelity.
Our main contributions are summarized as follows:
\begin{itemize}
\item[\textbullet] A Hierarchical Feature Modulation (HFM) strategy performs class-aware, spatially-differentiated alignment, overcoming semantically blind mixing for high distributional fidelity.

\item[\textbullet] A novel Hypergraph-based validation pipeline (HPE and SAP) moves beyond pixel-level confidence to ensure global anatomical plausibility, gating and refining structurally coherent pseudo-labels.

\item[\textbullet] Our integrated SHAPE framework establishes a robust UDA paradigm that significantly outperforms state-of-the-art methods via precise feature adaptation and high-fidelity pseudo-label supervision.
\end{itemize}

\section{Related Work}
\subsection{Self-Supervised Foundation Models}
Self-supervised learning (SSL) paradigms, including contrastive learning~\cite{Larochelle2020simclr,he2020momentum,Chen_2021_ICCV,radford2021learning} and masked image modeling~\cite{peng2022beit,he2022masked}, facilitate the pre-training of powerful Vision Transformers (ViTs) on unlabeled data. The resulting foundation models, such as DINOv3~\cite{simeoni2025dinov3}, CLIP~\cite{radford2021learning}, and SAM~\cite{kirillov2023segment}, have been pivotal in medical imaging~\cite{gui2024survey} by providing robust priors that enhance domain generalization. While these models offer a strong foundation, we contend that their direct application as feature extractors is insufficient for optimal adaptation. Consequently, we introduce a Hierarchical Feature Modulation (HFM) module designed to explicitly adapt these potent yet domain-biased priors to the specific characteristics of the target domain.

\subsection{Feature Alignment}
The evolution of feature alignment strategies is central to progress in UDA. Common strategies include adversarial alignment to enforce domain-invariance~\cite{wu2020dual,Hong2021UnsupervisedDAA,Lin2026DRLSTNet}, and global style alignment through statistical normalization like AdaIN~\cite{huang2017arbitrary} or spectral manipulation~\cite{xian2023unsupervised,yin2025ddfp}. Although effective for holistic appearance, these monolithic approaches are content-agnostic. To incorporate semantics, class-aware methods align per-class feature prototypes~\cite{yin2023class}. Interpolation strategies based on Mixup smooth the inter-domain transition by creating convex combinations of samples~\cite{Panfilov_2019_ICCV,kim2023bidirectional,cai2025style}. In contrast, our HFM performs content-aware, spatially-differentiated feature mixing. By dynamically applying either semantic interpolation or statistical alignment based on local patch content, HFM provides a granular, structure-preserving adaptation that overcomes the limitations of monolithic, semantically blind, or prototype-based methods.

\begin{figure*}[ht!]
  \centering
  \includegraphics[width=0.9\textwidth]{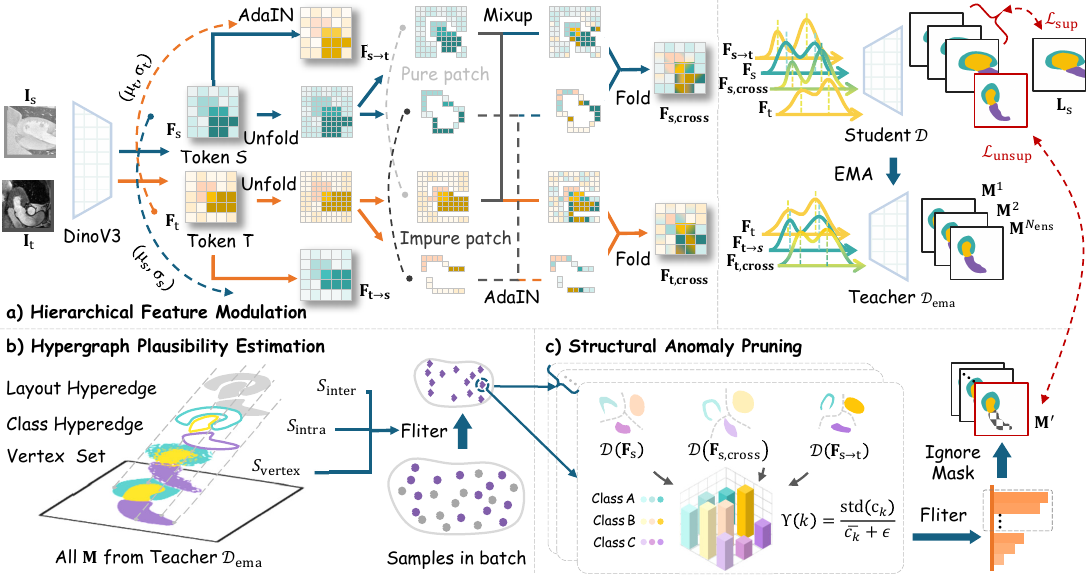}
  \caption{The pipeline of SHAPE. (a) Hierarchical Feature Modulation (HFM). (b) Hypergraph Plausibility Estimation (HPE). (c) Structural Anomaly Pruning (SAP).
  }
  \label{model}
\end{figure*}

\subsection{Structural Plausibility}
Incorporating anatomical priors is crucial for segmentation. Early methods imposed structural constraints through CRFs~\cite{chen2022end,koleilat2024medclip} or topology-aware losses~\cite{demir2023topology,katar2025att}. A better paradigm models relational context, where Graph Neural Networks (GNNs) capture pairwise object relations~\cite{xu2024g2vit} but are limited to binary interactions. Hypergraphs overcome this by representing higher-order relationships and have been applied to vision-based segmentation~\cite{jing2025multi,GUO2026111926,wang2025hypergraph}. Our work introduces the novel application of hypergraphs as a quality gate for pseudo-labels in UDA. We conceptualize predictions as structural hypergraphs to score the plausibility of intra-class shapes and inter-class layouts, creating a data-driven supervisory signal for self-training.

\section{Method}
As shown in \cref{model}, our proposed framework, SHAPE (\textbf{S}tructure-aware \textbf{H}ierarchical Unsupervised Domain \textbf{A}daptation with \textbf{P}lausibility \textbf{E}valuation), fundamentally reframes the UDA paradigm by shifting the objective from local pixel accuracy to global structural plausibility.  The framework begins with Hierarchical Feature Modulation (HFM), which constructs a structurally-aware feature space.
Although HFM can provide high-quality features, the subsequent challenge is to ensure that the resulting pseudo-labels are valid and accurate for efficient self-training.
We address this by introducing Hypergraph Plausibility Estimation (HPE), which models each prediction as a structural hypergraph to validate its global anatomical integrity. To achieve maximum fidelity, a final Structural Anomaly Pruning (SAP) stage then purges the remaining class-level instabilities. Through this cascade of feature adaptation and multi-level validation, SHAPE synthesizes a set of high-fidelity pseudo-labels to guide the model adaptation process.

\subsection{Hierarchical Feature Modulation}
Our approach is founded on a frozen DINOv3 ViT encoder \cite{simeoni2025dinov3}, which provides rich semantic and structural priors. For an input image $\mathbf{I}$, we extract the sequence of patch tokens from the final transformer block and reshape them into a dense feature map $\mathbf{F} =\Phi(\mathbf{I}) \in \mathbb{R}^{C \times H \times W}$, where $\Phi$ denotes the encoder. Our HFM module bridges the domain gap through a dual-granularity approach performing global style alignment and local, structure-aware feature mixing.

Globally, we use Adaptive Instance Normalization (AdaIN) \cite{huang2017arbitrary} to align textural properties between source $\mathbf{F}_\text{s}$ and target $\mathbf{F}_\text{t}$ features, computing the stylized map $\mathbf{F}_{\text{s} \to \text{t}}$ as:
\begin{equation}
\mathbf{F}_{\text{s} \to \text{t}} = \sigma(\mathbf{F}_\text{t}) \left( \frac{\mathbf{F}_\text{s} - \mu(\mathbf{F}_\text{s})}{\sigma(\mathbf{F}_\text{s}) + \epsilon} \right) + \mu(\mathbf{F}_\text{t}),
\end{equation}
where $\mu(\cdot)$ and $\sigma(\cdot)$ denote the channel-wise mean and standard deviation, and $\epsilon$ is a small constant added for numerical stability.

To perform a more nuanced local adaptation, we upsample the feature maps to a finer resolution, yielding denser grids of $N=4HW$ tokens. We extract these tokens and corresponding label sub-patches $\{\mathbf{m}_{\text{s}/\text{t}}^i\}_{i=1}^N$ through an unfolding operation $\mathcal{U}(\cdot)$. We then classify each token's content by computing a purity score for its aligned sub-patch:
\begin{equation}
\mathcal{P}(\mathbf{m}^i) = \frac{\max_{k \in \{0..K-1\}} \sum_{v \in \mathbf{m}^i} \mathbb{I}(v=k)}{|\mathbf{m}^i|},
\end{equation}
where $v$ is a pixel's class value, $K$ is the total number of classes, and $\mathbb{I}(\cdot)$ is the indicator function. Based on this score, we partition tokens into pure semantic cores ($\mathcal{T}_{\text{pure}}$) and impure structural boundaries ($\mathcal{T}_{\text{impure}}$). Specifically, a token is designated as pure if its purity score $\mathcal{P}(\mathbf{m}^i)$ exceeds a purity threshold $\tau_p$. This distinction guides a differentiated modulation strategy:
\begin{equation} \label{eq:hfm_combined}
\hat{\mathbf{f}}_\text{s}^i = 
\begin{cases} 
(1-\lambda)\mathbf{f}_\text{s}^i + \lambda \mathbf{f}_\text{t}^j & \text{if } i \in \mathcal{T}_{\text{s}, \text{pure}} \\
\sigma_{\text{mix}} \left( \frac{\mathbf{f}_\text{s}^i - \mu_\text{s}^{\text{impure}}}{\sigma_\text{s}^{\text{impure}} + \epsilon} \right) + \mu_{\text{mix}} & \text{if } i \in \mathcal{T}_{\text{s}, \text{impure}}
\end{cases}
\end{equation}
where $i$ and $j$ are indices for source and target tokens, respectively. For the pure case, $\lambda$ is a random mixing factor within the range [0, 1], and $\mathbf{f}_\text{t}^j$ is a target token of the same class as $\mathbf{f}_\text{s}^i$, selected from a pool sorted by proximity to the mean of the target class to prioritize mixing with representative exemplars. For the impure case, $\mu_\text{s}^{\text{impure}}$ and $\sigma_\text{s}^{\text{impure}}$ are the mean and standard deviation computed exclusively over the set of source boundary tokens $\{\mathbf{f}_\text{s}^i | i \in \mathcal{J}_{\text{s}, \text{impure}}\}$, and $\mu_{\text{mix}}, \sigma_{\text{mix}}$ are statistics interpolated from both source and target boundary tokens.

The modulated tokens $\{\hat{\mathbf{f}}_\text{s}^i\}$ are refolded to form the locally adapted map $\mathbf{F}_{\text{s}, \text{cross}}$. This hierarchical process is applied symmetrically to produce $\mathbf{F}_{\text{t} \to \text{s}}$ and $\mathbf{F}_{\text{t}, \text{cross}}$. All four feature maps are passed through the DINOv3 encoder's final layer normalization to ensure distributional consistency for the decoder.

While HFM furnishes structure-aware features, effective adaptation hinges on the anatomical plausibility of the pseudo-labels. We address this validation challenge next.

\subsection{Hypergraph Plausibility Estimation}
Reliable self-training requires anatomically plausible pseudo-labels, a quality beyond the reach of pixel-wise metrics. Therefore, we model each predicted segmentation map $\mathbf{M} \in \{0, ..., K-1\}^{H \times W}$ as a multi-level structural hypergraph $\mathcal{G} = (\mathcal{V}, \mathcal{E})$. The vertex set $\mathcal{V}$ consists of all foreground pixels, while the hyperedge set is the union of Class Hyperedges $\mathcal{E}_{\mathcal{C}} = \{e_k\}_{k=1}^{K-1}$, capturing intra-class shape, and a single Layout Hyperedge $e_l$, representing inter-class spatial arrangement. 

We first assess the vertex set $\mathcal{V}$ reliability through an aggregated score, $S_{\text{vertex}}$, which averages a pixel-wise weight $w_p$ over all foreground pixels:
\begin{equation}
S_{\text{vertex}}(\mathcal{G}) = \frac{1}{|\mathcal{V}|} \sum_{p \in \mathcal{V}} w_p,
\end{equation}
where the weight $w_p = (1 - \frac{H(\bar{\mathbf{M}}_p)}{\log K}) \cdot (1 - \frac{\text{JSD}(\{\mathbf{M}^{n}_p\})}{\log K})$ combines certainty (from mean entropy) and consistency (from JSD). The set of foreground pixels is $\mathcal{V}=\{p \mid \mathbf{M}_p>0\}$. The $\bar{\mathbf{M}}_p$ is the mean of $N_{\text{ens}}$ teacher predictions, and the JSD is defined as $\text{JSD}(\cdot) = H(\bar{\mathbf{M}}_p) - \frac{1}{N_{\text{ens}}}\sum_{n=1}^{N_{\text{ens}}} H(\mathbf{M}^{n}_p)$.

Beyond vertex quality, we evaluate structural coherence by scoring the hyperedges. For intra-class shape, encoded by the Class Hyperedges $\{e_k \in \mathcal{E}_{\mathcal{C}}\}$, the score $S_{\text{intra}}$ aggregates individual shape plausibility scores $S_{\phi,k}$ using a softmax-based weighting that heavily penalizes malformed outliers:
\begin{equation}
S_{\text{intra}}(\mathcal{G}) = \sum_{k=1}^{K-1} S_{\phi,k} \cdot \frac{\exp(-S_{\phi,k} / \tau)}{\sum_{j=1}^{K-1} \exp(-S_{\phi,j} / \tau)},
\end{equation}
where $S_{\phi,k} = \exp(-\lvert z_k \rvert)$ is derived from the Z-score $z_k = (\phi(e_k) - \mu_{\mathcal{B},\phi,k}) / (\sigma_{\mathcal{B},\phi,k} + \epsilon)$. The shape descriptor $\phi(e_k) = 4\pi \cdot \text{Area}(\mathbf{M}_k) / ((\text{Perimeter}(\mathbf{M}_k))^2 + \epsilon)$ is the isoperimetric ratio of the class mask $\mathbf{M}_k$, and the subscript $\mathcal{B}$ denotes numerically-stabilized statistics (mean $\mu$, std $\sigma$) computed over classes present in the current batch, and $\tau$ is a temperature parameter.

Analogously, the inter-class arrangement, encoded by the Layout Hyperedge $e_l$, is evaluated by the score $S_{\text{inter}}$, which aggregates plausibility scores derived from the relative direction cosines $\psi_{ij}$ between class centroids:
\begin{equation}
S_{\text{inter}}(\mathcal{G}) = \sum_{i,j=1}^{K-1} S_{\psi,ij} \cdot \frac{\exp(-S_{\psi,ij} / \tau)}{\sum_{u,v=1}^{K-1} \exp(-S_{\psi,uv} / \tau)},
\end{equation}
where$S_{\psi,ij}$ is computed in the same manner as $S_{\phi,k}$ using the descriptor $\psi_{ij}$ and its corresponding batch-level mean and standard deviation.

To derive a holistic measure of quality, the intra-class shape score ($S_{\text{intra}}$) and inter-class layout score ($S_{\text{inter}}$) are first linearly combined into a single structural plausibility metric. This metric then serves as a multiplicative gate for the vertex-level score ($S_{\text{vertex}}$), ensuring that predictions with high pixel confidence but poor anatomical structure are penalized:
\begin{equation}
S_{\text{final}} = S_{\text{vertex}}(\mathcal{G}) \cdot \left( \alpha S_{\text{intra}}(\mathcal{G}) + (1-\alpha) S_{\text{inter}}(\mathcal{G}) \right),
\end{equation}
where $\alpha$ is a fusion weight. Samples are selected for self-training only if $S_{\text{final}}$ exceeds a dynamic threshold determined by the top-$\rho$ percentile of scores accumulated within the current epoch.

\subsection{Structural Anomaly Pruning}
A prediction deemed globally plausible by HPE may still contain specific class-level artifacts, such as spurious regions that appear inconsistently or vary significantly in size across augmented views. Our pruning stage is designed to identify and remove these structurally anomalous classes.

Our method conceptualizes the stability of a predicted class $k$ by examining its structural signature across the ensemble of $N_{\text{ens}}$ teacher predictions. This signature is defined as the vector of pixel counts $\mathbf{c}_k = \langle C(\mathbf{M}^{1}, k), \dots, C(\mathbf{M}^{N_{\text{ens}}}, k) \rangle$, representing the class's size under differently modulated features. A robust anatomical structure should exhibit a stable signature with low variance, whereas a model hallucination is likely to manifest as a volatile signature. We quantify this volatility with a Structural Instability Score, $\Upsilon(k)$, defined as the coefficient of variation of the signature:
\begin{equation}
\Upsilon(k) = \frac{\text{std}(\mathbf{c}_k)}{\bar{c}_k + \epsilon},
\end{equation}
where $\bar{c}_k = \frac{1}{N_{\text{ens}}} \sum_{n=1}^{N_{\text{ens}}} c_{k,n}$ is the empirical mean of the pixel counts, $c_{k,m}$ denotes these counts of class $k$ in the $m$-th prediction, and $\text{std}(\cdot)$ computes the unbiased sample standard deviation. 

A class $k$ is deemed anomalous if its instability score exceeds a dynamic threshold $\theta_{\mathcal{A}}$, which is set to the $q$-th percentile of the instability scores from all significant foreground classes within the batch. This defines the set of anomalous classes:
\begin{equation}
\mathcal{K}_{\text{anom}} = \{k \in \{1, \dots, K-1\} \mid \Upsilon(k) > \theta_{\mathcal{A}} \}.
\end{equation}

Finally, the consensus pseudo-label map $\mathbf{M}$, which is generated from the teacher ensemble and has passed the HPE check, is pruned by masking all pixels belonging to the anomalous class set $\mathcal{K}_{\text{anom}}$. The refined map $\mathbf{M}'$ is thus defined for each pixel $p$ as:
\begin{equation}
(\mathbf{M}')_p = \begin{cases} (\mathbf{M})_p & \text{if } (\mathbf{M})_p \notin \mathcal{K}_{\text{anom}} \\ \text{ignore\_index} & \text{otherwise} \end{cases}.
\end{equation}

Through this cascade of structure-aware adaptation and validation, SHAPE synthesizes a set of high-fidelity pseudo-labels. These serve as the supervision signal for the student model, guiding its adaptation to the target domain with anatomically plausible targets.

\subsection{Overall Learning Objective}
The SHAPE framework is trained with a composite objective. For the source domain, the supervised loss $\mathcal{L}_{\text{sup}}$ trains the decoder $\mathcal{D}$ on source images with ground-truth labels $\mathbf{L}_\text{s}$. For domain robustness, it is computed as the average segmentation loss over the set of original and HFM-modulated features, defined as $\mathcal{F}_\text{s} = \{\mathbf{F}_\text{s}, \mathbf{F}_{\text{s} \to \text{t}}, \mathbf{F}_{\text{s}, \text{cross}}\}$:
\begin{equation}
\mathcal{L}_{\text{sup}} = \frac{1}{|\mathcal{F}_s|} \sum_{\mathbf{F}' \in \mathcal{F}_\text{s}} \mathcal{L}_{\text{seg}}(\mathcal{D}(\mathbf{F}'), \mathbf{L}_\text{s}).
\end{equation}

For the target domain, the unsupervised loss $\mathcal{L}_{\text{unsup}}$ is guided by high-fidelity pseudo-labels $\mathbf{M}'$. These are synthesized by our full validation pipeline (HPE and SAP), which processes a prediction ensemble generated by a teacher model, $\mathcal{D}_{\text{ema}}$, from modulated target features ($\mathbf{F}_\text{t}, \mathbf{F}_{\text{t} \to \text{s}}, \mathbf{F}_{\text{t}, \text{cross}}$). The loss supervises the student's predictions on the subset of samples $\mathcal{B}_{\text{sel}}$ that pass the plausibility check, weighted by pixel-wise certainty $w_p$:
\begin{equation}
\mathcal{L}_{\text{unsup}} = \frac{1}{|\mathcal{B}_{\text{sel}}|} \sum_{i \in \mathcal{B}_{\text{sel}}} \mathcal{L}_{\text{seg}}(\mathcal{D}(\mathbf{F}_\text{t}^i), (\mathbf{M}')^i, w_p^i).
\end{equation}
The total loss is $\mathcal{L}_{\text{total}} = \mathcal{L}_{\text{sup}} + \gamma_{\text{unsup}} \mathcal{L}_{\text{unsup}}$, where $\gamma_{\text{unsup}}$ is a ramp-up weight. The teacher model $\mathcal{D}_{\text{ema}}$ is an Exponential Moving Average (EMA)~\cite{tarvainen2017mean} of the student decoder, and $\mathcal{L}_{\text{seg}}$ is a standard segmentation loss, implemented as a combination of Dice~\cite{milletari2016v} and Focal loss~\cite{lin2020focal}.

\section{Experiments and Results}
\subsection{Experimental details}
\textbf{Datasets and metrics.} Our experiments are conducted on the cardiac dataset and the abdominal dataset. The cardiac dataset employs the MMWHS~\cite{zhuang2016multi} dataset, which comprises 20 3D CT scans and 20 3D MRI scans, with segmentation targets including the ascending aorta (AA), left atrium blood cavity (LAC), left ventricle blood cavity (LVC), and myocardium of the left ventricle (MYO). The abdominal dataset consists of 30 abdominal CT images from the MICCAI 2015 Multi-Atlas Abdomen Labeling Challenge~\cite{landman2015multi} and 20 T2SPIR MRI images from the ISBI 2019 CHAOS Challenge~\cite{kavur2021chaos}, with segmentation targets for the liver (LIV), right kidney (RK), left kidney (LK), and spleen (SPL). Each image is normalized to zero mean and unit variance, with affine transformations such as rotation and scaling applied. The performance is evaluated using the Dice score (DSC) and average surface distance (ASD) metrics.

\begin{table*}[h!]
    \centering
        \caption{Quantitative comparison of different methods on the cardiac dataset. The best values are highlighted in \textbf{bold}, and the second-best are \underline{underlined}.}
    \label{Cardiac-Merged}
\renewcommand\arraystretch{1.05}
\setlength{\tabcolsep}{0.9mm}
\fontsize{7}{8.4}\selectfont

\begin{tabular}{c|ccccc|ccccc|ccccc|ccccc}
  \specialrule{1.5pt}{0pt}{0pt} 
   \multirow{3}{*}{Method} &\multicolumn{10}{c|}{Cardiac MRI $\rightarrow$ Cardiac CT} & \multicolumn{10}{c}{Cardiac CT $\rightarrow$ Cardiac MRI} \\ 
  \cline{2-21}
& \multicolumn{5}{c|}{DSC(\%)$\uparrow$} & \multicolumn{5}{c|}{ASD(mm)$\downarrow$} & \multicolumn{5}{c|}{DSC(\%)$\uparrow$} & \multicolumn{5}{c}{ASD(mm)$\downarrow$} \\
  \cline{2-21}
  & AA & LAC & LVC & MYO & Avg & AA & LAC & LVC & MYO & Avg & AA & LAC & LVC & MYO & Avg & AA & LAC & LVC & MYO & Avg \\
  \hline
  Supervised & 91.28 & 92.49 & 95.56 & 94.16 & 93.37 & 2.16 & 2.43 & 2.56 & 1.3 & 2.11 & 82.22 & 83.21 & 90.46 & 81.74 & 84.41 & 2.23 & 2.29 & 2.73 & 2.35 & 2.4 \\ 
  W/o adaptation  & 68.29 & 61.41 & 18.24 & 35.71 & 45.91 & 50.21 & 22.14 & 52.16 & 27.56 & 38.02 & 36.56 & 46.49 & 49.23 & 15.35 & 36.91 & 35.46 & 15.9 & 16.54 & 24.32 & 23.06 \\ 
  \hline
  CycleGAN~\cite{zhu2017unpaired} & 63.29 & 72.50 & 45.98 & 50.73 & 58.13 & 40.21 & 15.32 & 14.85 & 14.74 & 21.28 & 34.57 & 65.08 & 75.13 & 59.57 & 58.59 & 21.24 & 13.27 & 7.25 & 15.51 & 14.32 \\
  AdaptSegNet~\cite{tsai2018learning} & 67.80 & 69.35 & 60.97 & 53.38 & 62.88 & 30.02 & 10.81 & 14.51 & 13.06 & 17.10 & 47.68 & 62.45 & 73.91 & 61.64 & 61.42 & 20.84 & 13.34 & 11.32 & 16.46 & 15.49 \\
  ADVENT~\cite{vu2019advent} & 73.77 & 68.46 & 61.03 & 57.21 & 65.12 & 17.31 & 16.96 & 18.88 & 15.83 & 17.25 & 35.01 & 64.03 & 58.71 & 52.80 & 52.64 & 33.65 & 16.38 & 22.31 & 24.02 & 24.09 \\
  SIFA~\cite{chen2020unsupervised} & 82.72 & 75.21 & 75.41 & 65.17 & 74.63 & 12.13 & 8.66 & 9.21 & 10.88 & 10.22 & 55.47 & 66.43 & 72.52 & 60.69 & 63.78 & 15.86 & 12.42 & 13.39 & 13.67 & 13.84 \\
  SASAN~\cite{tomar2021self} & 82.22 & 75.78 & 79.26 & 68.44 & 76.43 & 13.12 & 10.75 & 10.43 & 11.12 & 11.36 & 60.46 & 72.82 & 79.48 & 69.49 & 70.56 & 16.72 & 10.07 & 7.02 & 10.97 & 11.20 \\
  GenericSSL~\cite{wang2024towards} & 82.02 & 77.18 & 84.28 & 67.65 & 77.78 & \underline{3.23} & 8.72 & 6.14 & 8.45 & 6.64 & 63.35 & 72.77 & 84.04 & 72.38 & 73.14 & 16.06 & 11.14 & 5.96 & 7.33 & 10.12 \\
  UPL-SFDA~\cite{wu2023upl} & \underline{85.41} & 74.78 & 85.09 & 71.44 & 79.18 & 8.01 & 7.74 & 7.60 & 10.44 & 8.45 & \underline{67.71} & 75.18 & 80.59 & 72.77 & 74.06 & 13.95 & 9.94 & 7.57 & 7.27 & 9.68 \\
  IPLC~\cite{zhang2024iplc} & \textbf{87.63} & 78.21 & 86.11 & 71.68 & 80.91 & 5.65 & 7.55 & 3.96 & 8.31 & 6.37 & 67.55 & 75.98 & \underline{88.94} & 71.81 & 76.07 & 11.48 & 11.61 & 5.14 & 7.52 & 8.94 \\
  IPLC+~\cite{zhang2025iplc+} & 63.69 & \underline{86.15} & 86.71 & 89.17 & 81.43 & 4.21 & 4.14 & 3.34 & 3.95 & 3.91 & 65.09 & \underline{77.56} & \textbf{88.95} & \underline{74.33} & \underline{76.48} & \underline{4.69} & \underline{8.48} & \textbf{2.88} & \underline{5.44} & \underline{5.37} \\
  DDFP~\cite{yin2025ddfp} & 72.03 & 85.30 & \underline{89.64} & \underline{90.86} & \underline{84.46} & 4.52 & \underline{3.88} & \underline{3.02} &\underline{ 3.72} & \underline{3.79} & 66.26 & 76.04 & 88.55 & 70.64 & 75.37 & 7.72 & 10.33 & \underline{3.31} & 12.94 & 8.58 \\
  \hline
  SHAPE & 79.58 & \textbf{92.18} & \textbf{94.53} & \textbf{94.03} & \textbf{90.08} & \textbf{2.73} & \textbf{3.26} & \textbf{2.72} & \textbf{1.76} & \textbf{2.62} & \textbf{70.25} & \textbf{79.11} & 86.08 & \textbf{78.59} & \textbf{78.51} & \textbf{4.48} & \textbf{5.11} & 3.97 & \textbf{5.26} & \textbf{4.70} \\
  \specialrule{1.5pt}{0pt}{0pt} 
\end{tabular}
\end{table*}

\begin{table*}[ht]
\centering
\caption{Quantitative comparison of different methods on the abdominal dataset. The best values are highlighted in \textbf{bold}, and the second-best are \underline{underlined}.}
\label{Abdominal-Merged}
\renewcommand\arraystretch{1.05}
\setlength{\tabcolsep}{0.9mm}
\fontsize{7}{8.4}\selectfont

\begin{tabular}{c|ccccc|ccccc|ccccc|ccccc}
  \specialrule{1.5pt}{0pt}{0pt} 
    \multirow{3}{*}{Method}& \multicolumn{10}{c|}{Abdominal MRI $\rightarrow$ Abdominal CT} & \multicolumn{10}{c}{Abdominal CT $\rightarrow$ Abdominal MRI} \\ 
  \cline{2-21}
& \multicolumn{5}{c|}{DSC(\%)$\uparrow$} & \multicolumn{5}{c|}{ASD(mm)$\downarrow$} & \multicolumn{5}{c|}{DSC(\%)$\uparrow$} & \multicolumn{5}{c}{ASD(mm)$\downarrow$} \\
  \cline{2-21}
  & LIV & RK & LK & SPL & Avg  & LIV & RK & LK & SPL & Avg  & LIV & RK & LK & SPL& Avg  & LIV & RK & LK & SPL & Avg \\
  \hline
Supervised & 90.59 & 93.78 & 91.82 & 92.43 & 92.16 & 2.45 & 1.16 & 0.96 & 1.85 & 1.61 & 89.51 & 90.35 & 89.26 & 91.63 & 90.19 & 2.18 & 1.83 & 1.16 & 2.24 & 1.85 \\ 
W/o adaptation  & 39.36 & 26.14 & 42.46 & 52.36 & 40.08  & 32.16 & 18.11 & 25.31 & 35.26 & 27.71 & 35.14 & 48.26 & 33.51 & 49.25 & 41.54 & 31.51 & 22.62 & 13.16 & 34.63 & 25.48 \\ 
\hline
CycleGAN~\cite{zhu2017unpaired}            & 81.94 & 83.12 & 81.38 & 80.88 & 81.83 & 12.63 & 8.51 & 5.19 & 16.06 & 10.60 & 84.31 & 84.37 & 79.48 & 81.06 & 82.31 & \underline{6.38} & 4.78 & 10.55 & 15.70 & 9.35 \\
AdaptSegNet~\cite{tsai2018learning}      & 80.07 & 84.45 & 82.17 & 83.03 & 82.43 & 9.97 & 5.68 & 6.19 & 19.22 & 10.27 & 82.34 & 86.55 & 75.92 & 90.00 & 83.70 & 7.94 & 10.15 & 7.29 & 4.40 & 7.45 \\
ADVENT~\cite{vu2019advent}              & 81.11 & 82.70 & 83.34 & 81.17 & 82.08 & 11.01 & 7.05 & 4.84 & 10.02 & 8.23 & 79.64 & 81.49 & 75.77 & 81.15 & 79.51 & 10.28 & 7.05 & 9.96 & 19.68 & 11.74 \\
SIFA~\cite{chen2020unsupervised}                     & 83.64 & 86.67 & 83.21 & 79.88 & 83.35 & 6.41 & 5.47 & 4.39 & 8.35 & 6.16 & 83.65 & 86.79 & 77.19 & 89.03 & 84.17 & 18.77 & 13.86 & 5.23 & 4.26 & 10.53 \\
SASAN~\cite{tomar2021self}                 & 82.13 & 83.76 & 85.99 & 80.51 & 83.10 & 19.88 & 8.56 & \underline{2.83} & 8.34 & 9.90 & 82.75 & 85.26 & 78.92 & \underline{90.36} & 84.32 & 6.50 & 14.52 & 7.09 & 3.44 & 7.89 \\
GenericSSL~\cite{wang2024towards}    & 84.98 & 84.08 & 84.37 & \underline{84.92} & 84.59 & 10.38 & 7.78 & 5.76 & 6.32 & 7.56 & 82.43 & 86.88 & 80.52 & 89.79 & 84.91 & 8.15 & 9.11 & 6.78 & \textbf{2.79} & 6.71 \\
UPL-SFDA~\cite{wu2023upl}           & 84.21 & 86.34 & \underline{87.58} & 82.13 & 85.07 & 18.82 & 4.75 & \textbf{2.51} & 9.25 & 8.83 & \underline{84.39} & 86.65 & 78.44 & \textbf{90.76} & 85.06 & 6.61 & 11.57 & 9.92 & 6.19 & 8.57 \\
IPLC~\cite{zhang2024iplc}                     & 84.68 & 88.18 & 84.56 & 84.52 & 85.49 & 10.61 & 7.27 & 6.74 & 5.98 & 7.65 & 82.30 & 88.11 & 81.31 & 87.64 & 84.84 & 11.72 & 3.06 & 6.19 & 3.78 & 6.19 \\
IPLC+~\cite{zhang2025iplc+} & 85.29 & \textbf{88.63} & 85.12 & 83.79 & \underline{85.71} & 6.56 & 2.89 & 6.57 & 4.56 & 5.15 & 80.80 & \underline{88.30} & 85.71 & 85.41 & 85.06 & 13.68 & \underline{2.41} & 2.90 & \underline{3.33} & 5.58 \\
  DDFP~\cite{yin2025ddfp} & \underline{86.40} & \underline{87.66} & \textbf{88.05} & 78.55 & 85.17 & \textbf{3.24} & \textbf{2.15} & 3.57 & \underline{3.60} & \underline{3.14} & 82.49 & 86.43 & \underline{87.15} & 89.01 & \underline{86.27} & 11.57 & 2.96 & \underline{2.76} & 3.39 & \underline{5.17} \\
\hline
SHAPE             & \textbf{88.26} & 86.99 & 83.37 & \textbf{91.28} & \textbf{87.48} & \underline{3.84} & \underline{2.53} & 3.41 & \textbf{1.98} & \textbf{2.94}  & \textbf{86.83} & \textbf{88.86} & \textbf{88.30} & 83.58 & \textbf{86.89}  & \textbf{3.69} & \textbf{2.14} & \textbf{1.86} & 3.55 & \textbf{2.81}  \\
  \specialrule{1.5pt}{0pt}{0pt} 
\end{tabular}
\end{table*}

\noindent\textbf{Implementation details.} Our PyTorch framework uses a pre-trained frozen DINOv3 ViT-S/16 encoder and a trainable UNet-style decoder. We train for 200 epochs on a single NVIDIA RTX 4090 with a batch size of $64$ using the AdamW optimizer. The initial learning rate is $1 \times 10^{-4}$ with a cosine annealing schedule, and a teacher model is updated via EMA with $0.9$ momentum. All input images are resized to $256 \times 256$, replicated to 3 channels, and normalized using DINO statistics. Our data augmentation pipeline includes random contrast adjustments, random zoom, and random affine transformations. Key hyperparameters for our method are the HFM purity threshold $\tau_p=1$, a plausibility fusion weight $\alpha=0.25$, a selection percentile $\rho$ starting at 0.1 with a sigmoid ramp-up, an anomaly threshold $\theta_{\mathcal{A}}$ at the 50th percentile of instability scores, and the unsupervised loss weight $\gamma_{\text{unsup}}=1$.

\subsection{Comparison with state-of-the-art methods}
We compared several state-of-the-art UDA methods, including alignment-based ones (CycleGAN~\cite{zhu2017unpaired}, AdaptSegNet~\cite{tsai2018learning}, ADVENT~\cite{vu2019advent}, SIFA~\cite{chen2020unsupervised}, SASAN~\cite{tomar2021self}, DDFP~\cite{yin2025ddfp}) and pseudo-labeling ones (GenericSSL~\cite{wang2024towards}, UPL-SFDA~\cite{wu2023upl}, IPLC~\cite{zhang2024iplc}, and IPLC+~\cite{zhang2025iplc+}).

\begin{figure*}[ht!]
  \centering
  \includegraphics[width=0.95\textwidth]{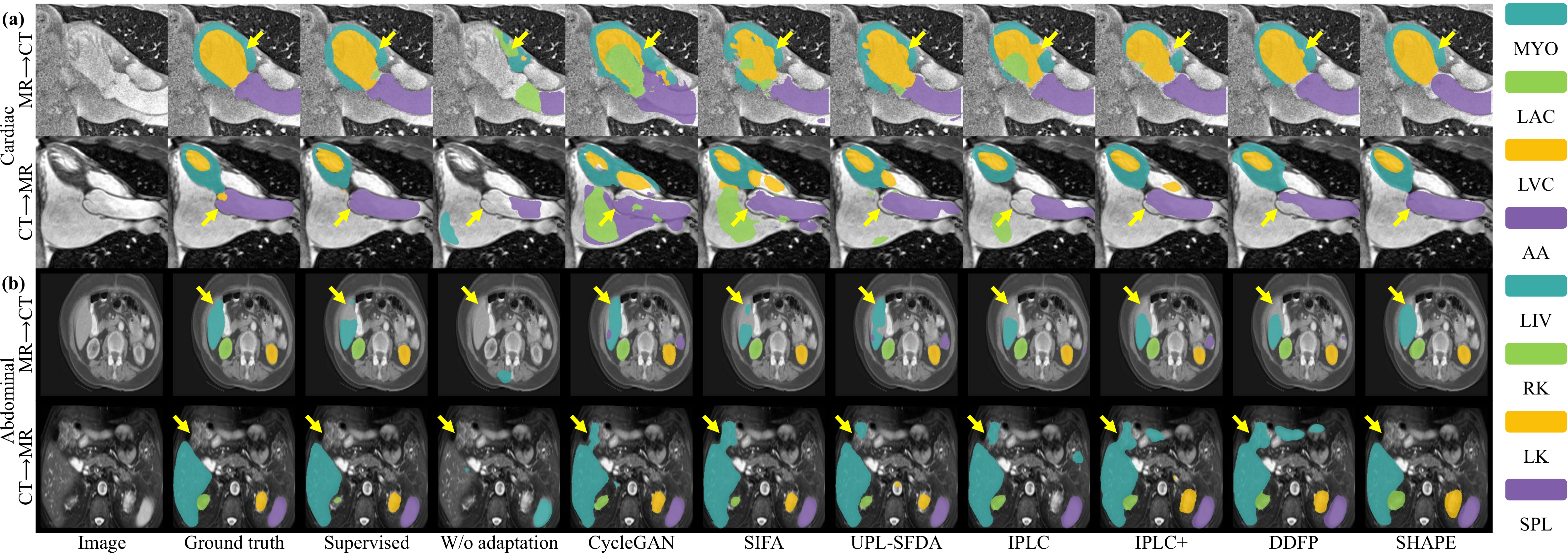}
  \caption{Qualitative results of SHAPE and typical methods. Yellow arrows indicate areas where SHAPE outperforms competing methods.}
  \label{result}
\end{figure*}

\begin{figure*}[ht!]
  \centering
  \includegraphics[width=0.95\textwidth]{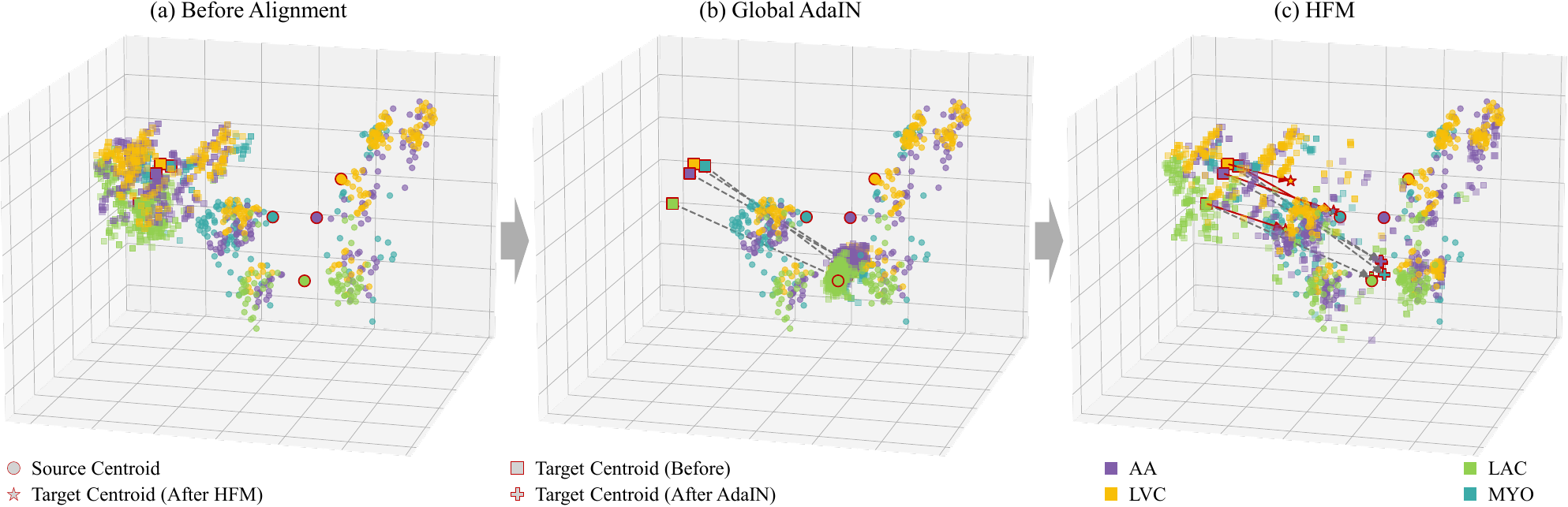}
  \caption{ (a) Feature distribution before adaptation. (b) Feature distribution after adaptation by AdaIN. (c) Feature distribution after adaptation by HFM.}
  \label{tsne}
\end{figure*}

\textbf{Quantitative comparison.} \Cref{Cardiac-Merged} and \cref{Abdominal-Merged} present the quantitative evaluation results for various UDA methods. The ``W/o adaptation'' results serve as the baseline lower bound, highlighting the significant domain gap between the source and target domains. For example, in the Cardiac MRI $\rightarrow$ Cardiac CT scenario, the average DSC drops to 45.91\% compared to the supervised upper bound. Similarly, the ASD increases to 38.02 mm, indicating poor segmentation quality and inaccurate boundary alignment when no adaptation is applied.

Notably, on the cardiac dataset, our SHAPE alleviates the performance degradation caused by domain shifts and outperforms all existing approaches across both domain adaptation scenarios. For example, in the Cardiac MRI $\rightarrow$ Cardiac CT scenario, SHAPE achieves a DSC of 90.08\%, marking a substantial 5.62\% improvement over the second-best method (DDFP~\cite{yin2025ddfp}) and narrowing the gap to the supervised upper bound to just 3.29\%.
Similarly, as shown in \cref{Abdominal-Merged}, the effectiveness of SHAPE is validated on the abdominal dataset, further demonstrating its capability. 

SHAPE's superior performance is driven by its synergistic pipeline. The HFM first generates domain-agnostic features through class-aware, spatially-differentiated alignment. These features produce high-quality initial pseudo-labels, which are then rigorously validated. HPE discards anatomically nonsensical predictions, while SAP purges the remaining class-level artifacts. This ensures self-training on high-fidelity labels, which directly improves segmentation performance.

\textbf{Qualitative comparison.} Visualizations of the final segmentation outputs in \cref{result} illustrate that our SHAPE produces fewer false predictions compared to other methods. To understand the underlying mechanism driving this improvement, we analyze the 3D t-SNE projection of patch-level features from the cardiac dataset in \cref{tsne}.
Initially, \cref{tsne}(a) reveals a clear domain gap, with source (circles) and target (squares) features occupying separate regions, and their respective class centroids are far apart. Global AdaIN, shown in \cref{tsne}(b), attempts a coarse alignment by pulling the target centroids ({crossed markers}) towards the source centroids. However, this monolithic transformation induces a severe distributional contraction, whereby target features from all classes are indiscriminately aggregated and lose their inter-class separability. This homogenization of feature representations is highly detrimental to a pixel-level segmentation task.
In stark contrast, \cref{tsne}(c) demonstrates HFM's structure-aware approach. The target class centroids ({stars}) are precisely aligned with their source counterparts, achieving superior domain adaptation. Crucially, this alignment occurs while preserving the intra-class distributional structure. The feature points of each class maintain their inherent variance and relative organization around their newly aligned centroids, avoiding the homogenization characteristic of AdaIN. This empirically validates that HFM successfully bridges the domain gap while retaining the fine-grained, discriminative feature structure essential for accurate segmentation.

\subsection{Ablation study}
\textbf{Effectiveness of Individual Components.} We first evaluate the contribution of each module by incrementally adding them to a strong baseline that already incorporates a DINOv3 backbone. The results, summarized in \cref{main_ablation}, demonstrate the efficacy of each component across two domain adaptation tasks. The baseline itself achieves a respectable DSC of 82.02\% on the MRI $\to$ CT task. Integrating our HFM provides the most significant individual performance improvement, improving the DSC by 3.65\% percentage points to 85.67\%. This underscores the critical importance of moving beyond global alignment to a class-aware, structure-preserving feature modulation strategy. Adding HPE alone also yields a consistent improvement, confirming that validating the anatomical plausibility of pseudo-labels is an effective strategy in its own right. The full SHAPE model, integrating all three modules, achieves the highest performance, reaching a DSC of 90.08\% on MRI $\to$ CT and 78.51\% on CT $\to$ MRI. The significant increase in performance when all components are active confirms that they are not merely additive but work synergistically to achieve the final result.

\begin{table}[h!]
\centering
\caption{Ablation study of the core components of SHAPE on the cardiac dataset. We evaluate the individual and combined contributions of our main modules over a strong baseline. The best results are highlighted in \textbf{bold}.}
\label{main_ablation}
\renewcommand\arraystretch{1.1}
\setlength{\tabcolsep}{1mm}
\fontsize{7}{8.4}\selectfont
\begin{tabular}{@{}l|ccc|cc|cc@{}}
\specialrule{1pt}{0pt}{0pt} 
\multicolumn{1}{c|}{\multirow{2}{*}{\textbf{Configuration}}} & \multicolumn{3}{c|}{\textbf{Modules}} &  \multicolumn{2}{c|}{MRI $\rightarrow$ CT} & \multicolumn{2}{c}{CT $\rightarrow$ MRI} \\ \cline{2-4} \cline{5-6} \cline{7-8}
\multicolumn{1}{c|}{} & HFM & HPE & SAP & DSC$\uparrow$ & ASD$\downarrow$ &  DSC$\uparrow$& ASD$\downarrow$  \\ \hline 
(a) Baseline & & & & 82.02 & 4.63 & 71.58 & 6.86\\ \hline
\multicolumn{6}{l}{\textit{Individual Component Contributions}} \\
(b) Baseline + HFM & \checkmark & & & 85.67 & 3.43 & 75.46 & 5.36 \\
(c) Baseline + HPE & & \checkmark & & 82.71 & 4.83 & 72.09 & 6.49\\ \hline
\multicolumn{6}{l}{\textit{Combined Component Contributions}} \\
(d) Baseline + HFM + HPE & \checkmark & \checkmark & & 85.80 & 3.24 & 75.81 & 5.17\\
(e) Baseline + HFM + SAP & \checkmark & & \checkmark& 86.03 & 3.02 & 76.23 & 5.13\\
(f) SHAPE (Full Model) & \checkmark & \checkmark & \checkmark & \textbf{90.08} & \textbf{2.62}& \textbf{78.51} & \textbf{4.70} \\ \hline  
\specialrule{1pt}{0pt}{0pt} 
\end{tabular}
\end{table}

\begin{figure}[h]
  \centering
  \includegraphics[width=0.5\textwidth]{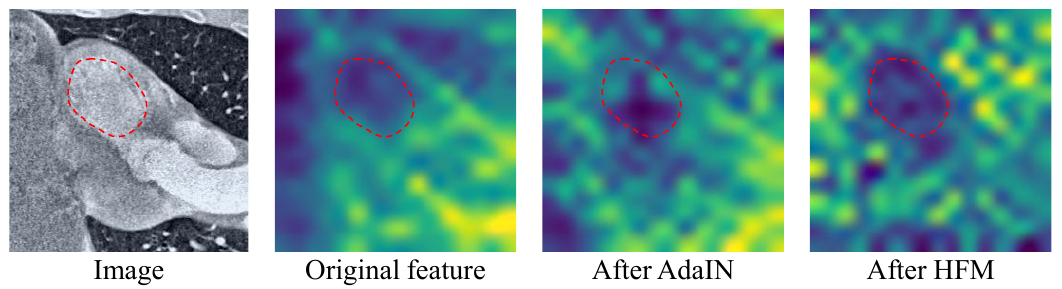}
  \caption{Comparison of the effectiveness of different feature modulation methods. Take the class inside the red dotted line as an example.}
  \label{feature}
\end{figure}

\noindent\textbf{Visual Analysis of Feature Modulation.} To visually substantiate why HFM is superior to global alignment, we visualize the feature maps of a target domain image in \cref{feature}. The ``Original feature" map demonstrates that the initial DINOv3 features already capture the anatomical structure within the region of interest (red dotted line). Applying global AdaIN, however, significantly disrupts this representation. The activations lose their structural coherence and become disorganized, failing to preserve the semantic boundaries of the anatomy. In stark contrast, the ``After HFM" map exhibits a remarkable semantic refinement. The activations become concentrated within the anatomical boundary, resulting in a cleaner and more discriminative feature representation. This visually demonstrates that HFM preserves feature quality through structure-aware alignment, whereas global AdaIN degrades it.

\begin{figure}[t]
  \centering
  \includegraphics[width=0.5\textwidth]{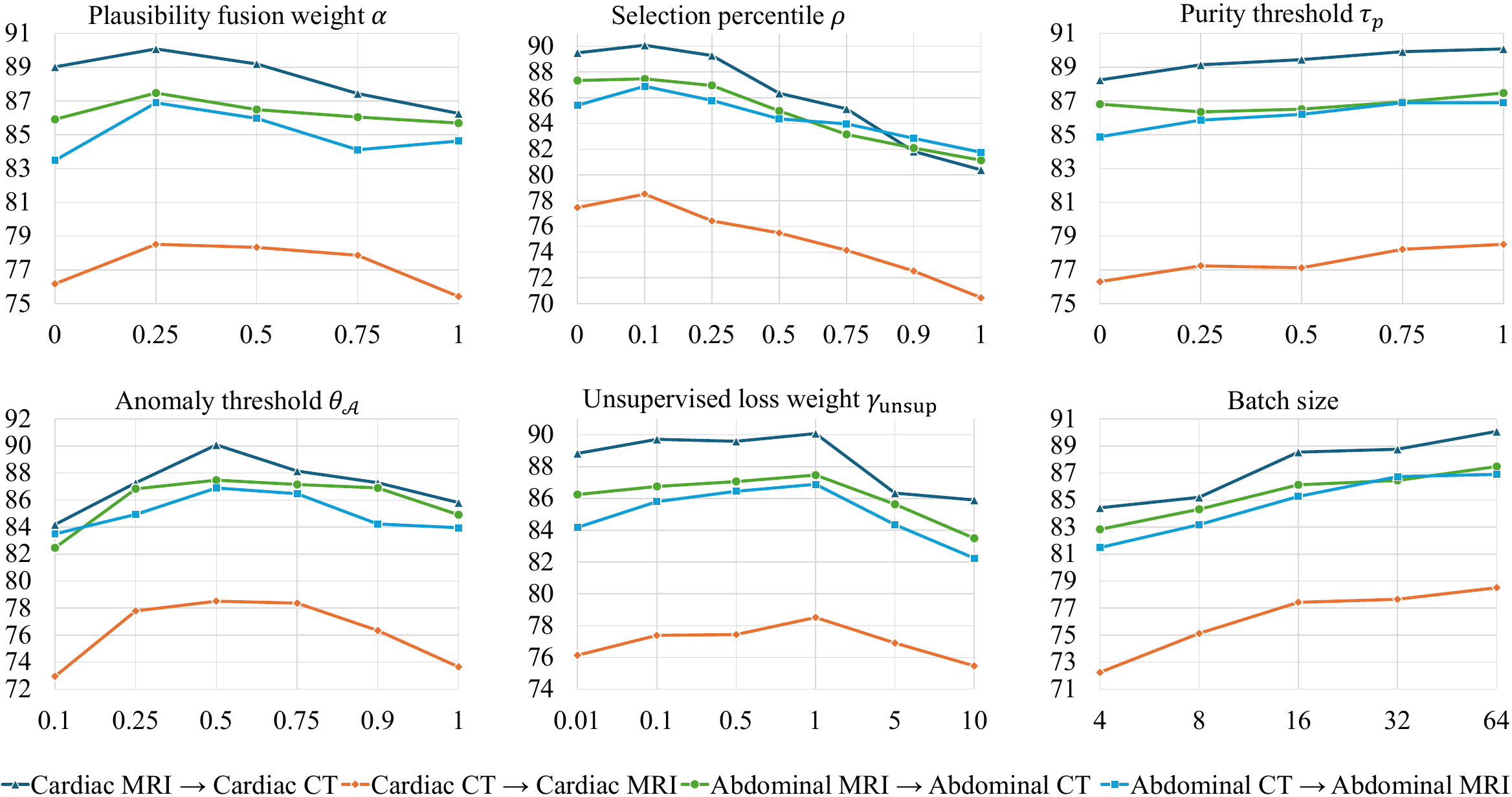}
  \caption{Hyperparametric sensitivity analysis of all adaptation segmentation tasks. The vertical axis is the mean Dice scores (\%) of all classes.}
  \label{para}
\end{figure}

\noindent\textbf{Hyperparameter Sensitivity Analysis.} To evaluate the robustness of SHAPE, we analyze its sensitivity to key hyperparameters in \cref{para}. Parameters governing trade-offs, such as the plausibility fusion weight $\alpha$ and anomaly threshold $\theta_{\mathcal{A}}$, exhibit clear optimal regions, validating our settings. Performance consistently improves with a stricter purity threshold $\tau_p$ and a larger batch size, as expected. Notably, while HPE benefits from larger batches for stable statistics, the framework remains effective at smaller sizes. This is because the SAP threshold $\theta_{\mathcal{A}}$ is derived from the distribution of every foreground class within the batch, providing a robust statistical basis even with fewer samples. Overall, the analysis confirms SHAPE's stability and practical applicability across a reasonable range of values.

\section{Conclusions}
In this work, we propose SHAPE, a novel framework for unsupervised domain adaptation in medical image segmentation. SHAPE first performs class-aware feature alignment through Hierarchical Feature Modulation (HFM) to overcome the limitations of semantically unaware adaptation. It then enforces the anatomical plausibility of pseudo-labels through a dual-validation pipeline, which uses Hypergraph Plausibility Estimation (HPE) to assess global coherence and Structural Anomaly Pruning (SAP) to purge local artifacts. Extensive experiments and ablation studies confirm that SHAPE significantly outperforms existing state-of-the-art methods in segmentation performance.


\section*{Acknowledgments}
This work was supported by the National Natural Science Foundation of China [Grant No. 62572401, No. 62222311, and 62322112], and the Key Research and Development Program of Shaanxi [Program No. 2025SF-YBXM-424].
{
    \small
    \bibliographystyle{ieeenat_fullname}
    \bibliography{main}
}


\end{document}